\input{psfig.sty}
\documentstyle[VLDB]{article}

\newcommand{\ca}{\cal A}
\newcommand{\cb}{\cal B}
\newcommand{\cc}{\cal C}

\newenvironment{myfigure}
{
\begin{figure*}[btp]
\begin{center}
\begin{tabular}[h]{cc}
\hspace*{160 mm} & \\ \hline
\end{tabular}
\vspace{2mm} \\
}
{
\begin{tabular}[h]{cc}
\hspace*{160 mm} & \\ \hline
\end{tabular}
\end{center}
\end{figure*}
}

\date{}
\title{\bf NeuroRule: A Connectionist Approach to Data Mining}
\author{
Hongjun Lu \hspace{0.5in}  Rudy Setiono \hspace{0.5in} Huan Liu \\
Department of Information Systems and Computer Science  \\
National University of Singapore    \\
\{luhj,rudys,liuh\}@iscs.nus.sg 
}

\institution{}

\begin{document}
\maketitle

\begin{abstract}

Classification, which involves finding rules that partition a given
data set into disjoint groups, is one class of data mining 
problems.  Approaches proposed so far for mining classification
rules for large databases are mainly decision tree based 
symbolic learning methods.
The connectionist approach based on neural networks 
has been thought not well suited for data mining.  One of the
major reasons cited is that knowledge generated by neural networks is
not explicitly represented in the form of rules suitable for verification
or interpretation by humans.  This paper examines this issue.
With our newly developed algorithms, rules which are similar to, or more 
concise than those generated by the symbolic methods
can be extracted from the neural networks.  The data mining process using
neural networks with the emphasis on rule extraction is described. 
Experimental results and comparison with previously published works
are presented.

\end{abstract}

\section{Introduction}

With the wide use of advanced database technology
developed during past decades, it is not difficult to efficiently 
store huge volume of data in computers and retrieve them
whenever needed.  Although the stored data are a valuable asset of 
an organization, most organizations may face the problem of {\em data rich but
knowledge poor \/} sooner or later.
This situation aroused the recent surge of research
interests in the area of data mining~\cite{Agra-etal92,Han-etal92,Agra-etal93}.

One of the data mining problems is {\em classification. \/}
Data items in databases, such as tuples in relational database systems
usually represent real world entities. 
The values of the attributes of a tuple represent the properties of 
the entity.  Classification is the process of finding the common 
properties among different entities and classifying them into $classes$.  
The results are often expressed in the form of rules -- the {\em
classification rules. \/}  By applying the rules,
entities represented by tuples can be easily classified
into different classes they belong to.
We can restate the problem formally defined  
by Agrawal {\em et al.\/} \cite{Agra-etal92} as follows.  
Let $A$ be a set of attributes  {$A_1, A_2, \ldots, A_n$} and
$dom(A_i)$ refer to the set of possible values for attribute $A_i$.  
Let $C$ be a set of classes {$c_1, c_2, \ldots, c_m$}.  
We are given a data set, {\em the training set \/}
whose members are $(n+1)$-tuples of the form 
($a_1, a_2, \ldots, a_n, c_k$) where $a_i \in  dom(A_i), (1 \leq i \leq 
n) $ 
and $c_k \in C ( 1 \leq k \leq m) $. Hence,
the class to which each tuple in the training set belongs
is known for supervised learning.
We are also given a second  large database of $(n+1)$-tuples, 
the testing set.
The classification problem is to obtain a set of rules $R$ using the
given  training data set.  By applying these rules to the testing set, the
rules can be checked whether they generalize well (measured by the
predictive accuracy). The rules that generalize well can be safely
applied to the application database with unknown classes to determine
each tuple's class.

This problem has been widely studied by  researchers in
the AI field~\cite{Weis-Kuli91}. 
It is recently re-examined by database researchers 
in the context of large database 
systems~\cite{Cerc-Tsuc93,Fraw-etal92,Piat92,Piat95,Math-etal93}. 
Two basic approaches
to the classification problems studied by AI researchers are the
symbolic approach and the connectionist approach.  The symbolic approach
is based on decision trees and the connectionist approach mainly uses
neural networks.  In general, neural networks give a lower classification
error rate than the decision trees but require longer learning 
time~\cite{Quin94,Shav-etal91,Russ-Norv95}.  
While both approaches have been well received by the AI community,
the general impression among the database community is that the
connectionist approach is not well suited for data mining.
The major criticisms include the following: 

\begin{enumerate}
\item  Neural networks learn the classification rules by multiple
       passes over the training data set so that the learning time, or
       the training time needed  for a neural network to obtain high 
       classification accuracy is usually long.
\item  A neural network is usually a layered graph  with the output
       of one node feeding into one or many other nodes in the next layer.
       The classification rules are buried in both the structure of
       the graph and the weights assigned to the links between the nodes.
       Articulating the classification rules becomes a difficult problem.
\item  For the same reason, available domain knowledge is rather difficult 
       to be incorporated to  a neural network.
\end{enumerate}

Among the above three major disadvantages of the connectionist approach, 
the articulating problem is the most urgent one to be solved for applying 
the technique to data mining.  Without explicit representation of
classification rules, 
it is very difficult to verify or interpret them.
More importantly, with explicit rules, tuples of a certain pattern can be
easily retrieved using a database query language.
Access methods such as indexing can be used or built for efficient retrieval 
as those rules usually involve only a small set of attributes.
This is especially
important for applications involving a large volume of data.  

In this paper, we present the results of our study on applying the neural
networks to mine classification rules for large databases with the
focus on articulating the classification rules represented by neural networks.
The contributions of our study include the following:

\begin{itemize}
\item
       Different from previous research work that
       excludes the connectionist approach entirely, we argue that
       the connectionist approach should have its position in data mining
       because of its merits such 
       as low classification error rates and  robustness to 
	noise~\cite{Quin94,Russ-Norv95}.

\item  With our newly developed algorithms, explicit classification rules 
       can be extracted from a neural network.  The rules extracted usually
       have a lower classification error rate than those generated by the
       decision tree based methods.  For a data set with  a strong relationship
       among attributes, the rules extracted are generally more concise.

\item A data mining system, NeuroRule, based on neural networks was
      developed. The system  successfully solved a number of classification
      problems in the literature.
\end{itemize}

To better suit large database applications, we also developed
algorithms for input data pre-processing and for fast neural network 
training to reduce the time needed to learn the classification
rules~\cite{Seti-Liu95a,Seti94b}. Limited by space, those algorithms are not 
presented in this paper.
     
The remainder of the paper is organized as follows.  
Section 2  gives a discussion on using the connectionist approach 
to learn classification rules.
Section 3 describes our algorithms to extract classification rules
from a neural network. Section 4 presents some experimental results obtained
and a comparison with previously published results.
Finally a conclusion is given in Section 5.

\section{Mining classification rules  using neural networks }
Artificial neural networks are densely interconnected networks of simple
computational elements, $neurons$. There exist many different network 
topologies \cite{Hert-etal91}. 
Among them, the {\em multi-layer perceptron \/} is 
especially useful for implementing a classification function.  
Figure~\ref{fig:nn} shows a three layer feedforward network.
It consists of an input layer, a hidden layer and an output
layer.  A node (neuron) in the network has a number of inputs
and a single output.
For example, a neuron $H_j$ in the hidden layer has $x_1^i, x_2^i, \ldots, 
x_n^i$ as its input and $\alpha^j$ as its output.
The input links of
$H_j$ has weights $w_1^j, w_2^j, \ldots, w_n^j$.
A node computes its output, the {\em activation value} by
summing up its weighted inputs, subtracting a threshold, and
passing the result to a non-linear function $f$, the 
{\em activation function. \/}
Outputs from neurons in one layer are fed as inputs to neurons in the next
layer. In this manner, when an input tuple is applied to the input layer,
an output tuple is obtained at the output layer.
For a well trained network which represents the classification
function,  if tuple ($x_1, x_2, \ldots, x_n$) is applied to the
input layer of the network, the output tuple, ($c_1, c_2, \ldots,
c_m$) should be obtained where $c_i$ has value 1 if the input tuple belongs 
to class $c_i$ and 0 otherwise.

Our approach that uses neural networks to mine classification rules
consists of three steps:
\begin{enumerate}
\item {\em Network training \/} \\
      A three layer neural network is  trained in this step.
      The training phase aims to find the best set of weights for the
      network which allow the network to classify input tuples with a
      satisfactory level of accuracy. An initial set of weights are
      chosen randomly in the interval [-1,1]. Updating these weights
      is normally done by using informations involving the gradient
      of an error function.
      This phase is terminated when the norm of the gradient of the error
      function falls below a prespecified value.

\item {\em Network pruning} \\
      The network obtained from the training phase is fully connected
      and could have too many links and sometimes too many nodes as well.
      It is impossible to extract concise rules which are meaningful to users 
      and can be used to form database queries from such a network.
      The pruning phase aims at removing redundant
      links and nodes without increasing the classification error rate 
      of the network. A smaller
      number of nodes and links left in the network after pruning 
      provide for extracting consise and comprehensible rules that describe 
      the classification function.

\item {\em Rule extraction} \\
      This phase extracts the classification rules from the pruned network.
      The rules generated are in the form of 
      ``if $(a_1 \theta v_1)$ {\em and \/} $(x_2 \theta v_2)$ 
      {\em and }  $\ldots$ {\em and \/} $(x_n \theta v_n)$ 
       then $C_j$'' 
      where $a_i$'s are the attributes of an input tuple,
      $v_i$'s are constants, $\theta$'s are relational operators 
      ($=, \leq, \geq, <>$),
      and $C_j$ is one of the class labels.  It is expected that the rules
      are concise enough for human verification and are easily applicable
      to large databases.
\end{enumerate}

In this section, we will briefly discuss the first two phase.  The third
phase, rule extraction phase will be discussed in the next section.

\subsection{Network training}

\begin{figure}[h]
\vspace*{5mm}
{\psfig {figure=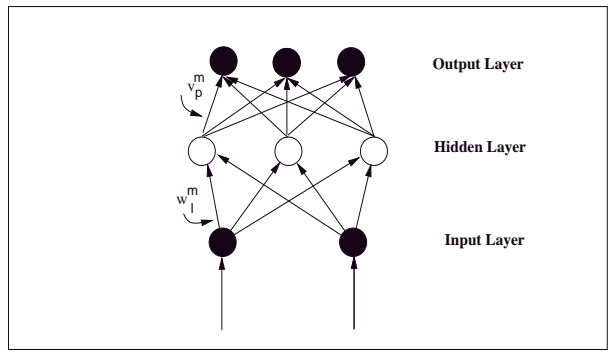,width=8cm,height=46mm}}   
\caption{A three layer feedforward neural network.}
\label{fig:nn}
\end{figure}

Assume that input tuples in an $n$-dimensional space
are to be classified into three disjoint classes  
$\ca, \cb,$ and $\cc$. 
We construct a  network as shown in Figure~\ref{fig:nn}
which consists of three layers.
The number of nodes in the input layer corresponds to the dimensionality of
the input tuples. The number of nodes in the
output layer equals to the number of classes to be classified,
which is three in this example.
The network is trained with target values equal to $\{1,0,0\}$
for all patterns in set $\ca$, $\{0,1,0\}$
for all patterns in $\cb$, and $\{0,0,1\}$ for all tuples in $\cc$.
An input tuples will be classified as a member of the class $\ca, \cb$ or 
$\cc$ if the  largest  activation value is obtained by the first, second 
or third output node, respectively.

There is still no clear cut rule to determine the number of hidden nodes to
be included in the network.
Too many hidden nodes may lead to overfitting of
the data and poor generalization,
while too few hidden nodes may not give rise to a network that learns the data.
Two different approaches have been proposed to overcome the problem of
determining the optimal number of hidden nodes required by a
neural network to solve a given problem.
The first approach begins with a minimal network and adds more hidden 
nodes only
when they are needed to improve the learning capability of the 
network~\cite{Ash89,Hiro-etal91,Seti94b}.
The second approach begins with an oversized network and then prunes redundant
hidden nodes and connections between the layers of the network.
We adopt the second approach since we are interested in finding a network
with a small number of hidden nodes as well as the fewest number of input
nodes. An input node with no connection to any of the hidden nodes after
pruning plays no role in the outcome of classification process and hence can be
removed from the network.
\nopagebreak

The activation value of a node in the hidden layer is 
computed by passing the weighted sum of input values to a non-linear
activation function.  
Let $w^{m}_{\ell}$ be the weights for the connections from input node $\ell$ to
hidden node $m$. 
Given an input pattern $x^{i}, i \in \{1,2,\ldots, k\}$,
where $k$ is the number of tuples in the data set,
the activation value of the $m$-th hidden node is
\[  \alpha^{m} = f\left(
   \sum_{\ell = 1}^{n} \left( x^{i}_{\ell} w^{m}_{\ell} \right) - \tau^{m}
   \right), \]
where $f(.)$ is an activation function. 
In our study, we use the hyperbolic tangent function  
\[ f(x) := \delta(x) = (e^{x} - e^{-x})/(e^{x} + e^{-x})\]
as the activation function for the hidden nodes, 
which makes the range of activation values of the hidden nodes [-1, 1]. 

Once the activation values of all the hidden nodes have been computed, the
$p$-th output of the network for input tuple $x^{i}$ is computed as
\[ S^{i}_{p} =
   \sigma\left( \sum_{m=1}^{h} \alpha^m v^m_p \right), \]
where $v^{m}_{p}$ is the weight of the connection between hidden node
$m$ and output node $p$ and
$h$ is the number of hidden nodes in the network.
The activation function used here is the sigmoid function,
\[ \sigma(x) = 1/(1 + e^{-x}), \]
which yields activation values of the output nodes in the range  [0, 1].


A tuple will be correctly classified if the following condition is satisfied
\begin{equation} \label{error} 
\max_{p}  |e^{i}_p| = \max_p |S^{i}_p - t^{i}_p| \le \eta_{1},\end{equation}
where $t^{i}_{p} = 0$, except for $t^{i}_{1} = 1$ if $
x^{i} \in \ca$, $t^{i}_{2} = 1$ if  $x^{i} \in \cb$, and 
$t^{i}_{3} = 1$ if $ x^{i} \in \cc$, and $\eta_{1}$ is 
a small positive number less than 0.5.
The ultimate objective of the training phase is to obtain a set of weights
that make the network classify the input tuples correctly.
To measure the classification error, an error function is needed so that
the training process becomes a process to adjust the weights
($w, v$) to minimize this function.  Furthermore,
to facilitate the pruning phase, it is desired to have many weights
with very small values so that they can be set to zero.
This is 
achieved by adding a penalty term to the error function. 

In our training algorithm, the cross entropy function 
\begin{equation} \label{eq:entropy}
E(w,v) = - \sum_{i=1}^{k} \sum_{p=1}^{o}
\left( t^i_p \log S^{i}_p + (1 - t^{i}_p) \log (1 - S^{i}_p) \right)
\end{equation}
is used as the error function.
In this example, $o$ equals to 3 since we have 3 different classes.
The cross entropy function is chosen
because faster convergence can be achieved by minimizing
this function instead of the widely used sum of
squared error function~\cite{Ooye-Nien92}.

The penalty term $P(w, v)$ we used is
\begin{equation} \label{new}
\footnotesize
\epsilon_{1} \left(
\sum_{m=1}^{h} \sum_{\ell=1}^{n}
\frac{\beta (w_{\ell}^{m})^{2}}{1 + \beta(w_{\ell}^{m})^{2}}
+ \sum_{m=1}^{h}\sum_{p=1}^{o} 
\frac{\beta ( v^{m}_p)^{2}} {1 + \beta (v^{m}_p)^{2} }\right) +
\normalsize
\end{equation}
\[ 
\footnotesize
\epsilon_{2} \left(
\sum_{m=1}^{h} \sum_{\ell=1}^{n}\left(w_{\ell}^{m}\right)^{2}
+\sum_{m=1}^{h} \sum_{p=1}^{o} \left(v^{m}_p\right)^{2}\right),
\normalsize
\]
where $\epsilon_{1}$ and $\epsilon_{2} $ are two positive 
weight decay parameters. 
Their values reflect the relative importance of the accuracy of the
network versus its complexity.
With larger values of these two parameters
more weights may be removed later from the network at the cost of a decrease in
its accuracy.  

The training phase starts with an initial
set of weights $(w,v)^{(0)}$ and iteratively updates the weights
to minimize $E(w,v) + P(w,v)$.
Any unconstrained minimization algorithm can be used for this purpose.
In particular, the gradient descent method has been the most widely used
in the training algorithm known as the backpropagation algorithm.
A number of   alternative  algorithms for neural network training
have been proposed~\cite{Batt92}. To 
reduce the network training time, which is very important in
the data mining as the data set is usually large,
we employed a variant of the quasi-Newton algorithm~\cite{Watr87}, the
BFGS method.
This algorithm has a superlinear convergence rate, as opposed to the linear
rate of the gradient descent method.
Details of the BFGS algorithm can be found in~\cite{Denn-Schn83,Shan-Phua76}.

The network training is terminated when a local minimum of the
function $E(w,v) + P(w,v)$ has been reached, that is when the gradient of
the function is sufficiently small.

\subsection{Network pruning}

A fully connected network is obtained at the end of the training process.
There are usually a large number of links in the
network. With $n$ input nodes, $h$ hidden nodes, and $m$ output nodes,
there are $h (m+n)$ links.  It is very difficult
to articulate such a network.  The network pruning phase aims at
removing some of the links without affecting the classification accuracy
of the network.

It can be shown that~\cite{Seti94a}
if a network is fully trained to correctly
classify an input tuple, $x^i$, with the condition (1) satisfied
we can set $w^{m}_{\ell}$ to zero
without deteriorating the overall accuracy  of the network if
the product $|v^{m} w^{m}_{\ell}|$ is sufficiently small.
If $\max_{p} |v^{m}_p w^{m}_{\ell}| \le 4 \eta_2$ and
the sum $(\eta_{1} + \eta_{2})$ is less than 0.5, 
then the network can still classify $x^i$ correctly.
Similarly, if $\max_{p} |v^m_p| \le 4 \eta_2$, then $v^m_p$ can 
be removed from the network.

\begin{figure}
\begin{tabular}{cc}
\hspace{72 mm} & \\ \hline \\
\end{tabular}
\vspace*{2mm} 
{\bf Neural network pruning algorithm (NP)}
\begin{enumerate}
\item  Let $\eta_{1}$ and $\eta_{2}$ be positive scalars such that
       $\eta_{1} +  \eta_{2} < 0.5$.
\item  Pick a fully connected network. Train this network until a 
       predetermined accuracy rate is achieved and for each correctly
       classified pattern the condition (\ref{error}) is satisfied.
       Let $(w,v)$ be the weights of this network.
\item  For each $w^{m}_{\ell}$, if 
       \begin{equation}\label{a}
	\max_{p} |v^{m}_p \times w_{\ell}^{m}| \le 4 \eta_{2}, \end{equation}
       then remove $w^{m}_{\ell}$ from the network
\item  For each $v^{m}_{p}$, if
       \begin{equation}\label{b}
         |v^{m}_p| \le 4 \eta_{2}, \end{equation}
       then remove $v^{m}_p$ from the network
\item  If no weight satisfies condition (\ref{a}) or condition (\ref{b}), then
       remove $w^{m}_{\ell}$ with the smallest product
	 $\max_{p} |v^{m}_p \times w_{\ell}^{m}|$.
\item  Retrain the network. If accuracy of the network falls
       below an acceptable level, then stop. Otherwise, go to Step 3.
\end{enumerate}
\caption {Neural network pruning algorithm }
\label{fig:NP}
\begin{tabular}[h]{cc}
\hspace*{72 mm} & \\ \hline
\end{tabular}
\end{figure}

Our pruning algorithm based on this result is shown in Figure~\ref{fig:NP}.
The two conditions (\ref{a}) and (\ref{b}) for pruning depend on the
magnitude of the weights for connections  between input nodes and hidden
nodes and between hidden nodes and output nodes. It is imperative that
during training these weights be prevented from getting too large.
At the same time, small weights should be encouraged to decay rapidly to zero.
By using penalty function (\ref{new}), we can achieve both.

\subsection{An example}

We have chosen to use a function described in~\cite{Agra-etal93} 
as an example to show how a neural network can be trained and
pruned for solving a classification problem.
The input tuple consists of nine attributes defined in Table~\ref{tab:desc}.
Ten classification problems are given in~\cite{Agra-etal93}.
Limited by space, we will present and discuss a few functions and the
experimental results.
\noindent
\begin{table*}[bt]
\begin{center}
\caption {Attributes of the test data adapted from Agrawal et al.[2]}
\label{tab:desc}
\vspace*{2mm}
\begin{tabular}{l l l} \hline \hline
\multicolumn{1}{c} {Attribute} & \multicolumn{1}{c} {Description}
& \multicolumn{1}{c} {Value} \\ \hline
 salary    &  salary     & uniformly distributed from 20,000 to 150,000\\
commission & commission  & if salary $\geq$ 75000 $\rightarrow$ commission = 0\\
           &             & else uniformly distributed from 10000 to 75000.\\
 age      &    age      & uniformly distributed from 20 to 80.\\
 elevel   & education level & uniformly distributed from $[0,1,\ldots, 4]$.\\
 car      & make of the car & uniformly distributed from $[1,2, \ldots 20]$.\\
 zipcode  & zip code of the town & uniformly chosen from  9 available zipcodes.\\
 hvalue   & value of the house & uniformly distributed from 0.5$k$10000 to 1.5$k$1000000 \\
          &                    & where $k \in \{0 \ldots 9\}$ depends on zipcode.\\
hyears    & years house owned & uniformly distributed from $[1, 2, \ldots, 30]$. \\
loan      &  total amount of loan & uniformly distributed from 1 to 500000.\\
\hline
\end{tabular}
\end{center}
\end{table*}

Function 2 classifies a tuple in Group A if 
\[
(({\bf age} < 40) \wedge (50000 \leq {\bf salary} \leq 100000)) \vee 
\]
\[
((40 \leq {\bf age} < 60) \wedge (75000 \leq {\bf salary} \leq 125000)) \vee 
\]
\[
(({\bf age} \geq 60) \wedge (25000 \leq {\bf salary} \leq 75000)).
\]
Otherwise, the tuple is classified in Group B.

The training data set consisted of 1000 tuples.  The values of the
attributes of each tuple were generated randomly according to the 
distributions given in Table~\ref{tab:desc}. Following Agrawal 
{\em et al. \/}~\cite{Agra-etal93},
we also included  a perturbation factor as one of the parameters of the random
data generator. This perturbation factor was set at 5 percent.
For each tuple, a class label was determined
according to the rules that define the function above.

To facilitate the rule extraction in
the later phase, the values of the numeric attributes were discretized.
Each of the six attributes with
numeric values was discretized by dividing its range into subintervals.
The attribute {\it salary} for example, which was uniformly distributed 
from 25000 to 150000 was divided into 6 subintervals: subinterval 1 contained
all salary values that were strictly less than 25000, subinterval 2
contained those greater than or equal to 25000 and
strictly less than 50000, etc.
The {\it thermometer} coding scheme was then employed to get the binary 
representations of these intervals for inputs to the neural network.
Hence, a salary value less that 25000 was coded as $\{ 000001 \}$,
a salary value in the interval $[25000,50000)$ was coded as
$\{ 000011 \}$, etc.
The second attribute {\it commission} was similarly coded.
The interval from 10000 to 75000 was divided into 7 subintervals, each 
having a width of
10000 except for the last one, $[70000,75000]$.
Zero commission was coded by all zero values for the seven inputs.
The coding scheme for the other attributes are given in 
Table~\ref{tab:binary}.

\begin{table}[hbt]
\begin{center}
\caption{Binarization of the attribute values}
\label{tab:binary}
\vspace*{2mm}
\begin{tabular}{l c c} \hline \hline
\multicolumn{1}{c} {Attribute} & \multicolumn{1}{c} {Input number}
& \multicolumn{1}{c} {Interval width}\\ \hline
salary    & ${\cal I}_{1}$ - ${\cal I}_6$    & 25000 \\
commission& ${\cal I}_{7}$ - ${\cal I}_{13}$ & 10000 \\
age       & ${\cal I}_{14}$ - ${\cal I}_{19}$ & 10 \\
elevel    & ${\cal I}_{20}$ - ${\cal I}_{23}$ & -  \\
car       & ${\cal I}_{24}$ - ${\cal I}_{43}$ & -  \\
zipcode   & ${\cal I}_{44}$ - ${\cal I}_{52}$ & -  \\
hvalue    & ${\cal I}_{53}$ - ${\cal I}_{66}$ & 100000  \\
hyears    & ${\cal I}_{67}$ - ${\cal I}_{76}$ & 3  \\
loan      & ${\cal I}_{77}$ - ${\cal I}_{86}$ & 50000  \\ \hline
\end{tabular}
\end{center}
\end{table}
With this coding scheme, we had a total of 86 binary inputs.
The 87th input was added to the network to incorporate the 
bias or threshold in each of the hidden node. The input value to this input was
set to one.  Therefore the input layer of the initial network
consisted of 87 input nodes.
Two nodes were used at the output layer.
The target output of the network was $\{1,0\}$ if the
tuple belonged to Group $A$, and $\{0,1\}$ otherwise.
The number of the hidden  nodes was initially set as four.

There were a total of 386 links in the network. The weights for these
links were given initial values that were randomly generated in the 
interval [-1,1].
The network was trained until a local minimum point of the error function
had been reached.

The fully connected trained network was then pruned by the pruning algorithm
described in Section 2.2.
We continued removing connections from the neural network as long as the
accuracy of the network was still higher than 90 \%.

\begin{figure}
\vspace*{5mm}
 {\psfig {figure=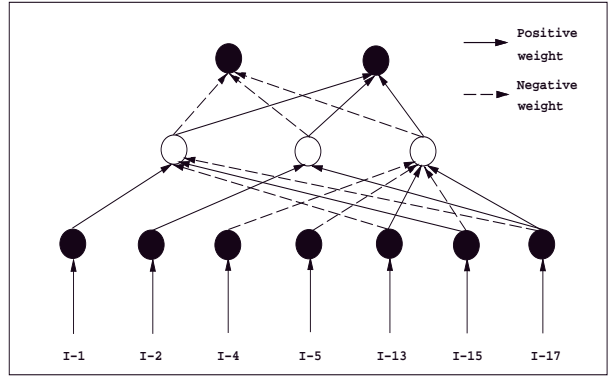,width=8cm,height=5cm}}   
\caption{Pruned network for Function 2. Its accuracy rate 
on the 1000 training samples is 96.30 \% and it contains only 17 connections.}
\label{fig:prun_nn}
\vspace*{5mm}
\end{figure}

Figure~\ref{fig:prun_nn} shows the pruned network.
Of the 386 links in the original network, only 17 remained in the
pruned network. One of the four hidden nodes was removed.
A small number of links from the input nodes to the hidden nodes
made it possible to extract compact rules with the same
accuracy level as the neural network.

\section{Extracting rules from a neural network}

Network pruning results in a relatively simple network.
In the example shown in the last section, the pruned network has only
7 input nodes, 3 hidden nodes, and 2 output nodes.  The number of
links is 17.  However, it is still very difficult to articulate
the network, i.e., find the explicit relationship between the input
tuples and the output tuples.  Research work in this area
has been reported~\cite{Towe-Shav93,Fu94}.  However, to
our best knowledge, there is no method available in the
literature that can extract explicit and concise rules as the algorithm
we will describe in this section.

\subsection{Rule extracting algorithm}

A number of reasons contribute to the difficulty of extracting rules
from a pruned network.  First, even with a pruned network, the links
may be still too many to express the relationship between an input tuple
and its class label in the form of {\em if $\ldots$ then $\cdots$ \/} rules.
If a node has $n$ input links with binary values, there could be as many as
$2^n$ distinct input patterns.  The rules could be quite lengthy or complex 
even
with a small $n$, say 7. Second, the activation values of a hidden node
could be anywhere in the range [-1,1] depending on the input tuple. With
a large number of testing data, the activation values are virtually continuous.
It is rather difficult to derive the explicit relationship between the
activation values of the hidden nodes and the output values of a node
in the output layer.

\begin{figure}
\begin{tabular}{cc}
\hspace{72 mm} & \\ \hline
\end{tabular}
\vspace*{2mm} 
{\bf Rule extraction algorithm (RX)}
\begin{enumerate}
\item Activation value discretization via clustering:
\begin{enumerate}
\item Let $\epsilon \in (0,1)$. Let $D$ be the number of discrete activation
      values in the hidden node.
      Let $\delta_{1}$ be the activation value in the hidden node for the
      first pattern in the training set.
      Let $H(1) = \delta_1, count(1) = 1, sum(1) = \delta_1$ and set $D = 1$.
\item For all patterns $i = 2,3, \ldots k$ in the training set:
      \begin{itemize}
      \item Let $\delta$ be its activation value.
      \item If there exists an index $\overline{j}$
	    such that
            \begin{eqnarray*}
	| \delta - H(\overline{j}) | &=& \min_{j \in \{1,2,\ldots, D\}}
	| \delta - H(j)| \;\; \\
        \mbox{and} | \delta - H(\overline{j}) | &\le& \epsilon,  
	    \end{eqnarray*}
	    then set $count(\overline{j})$ := $count(\overline{j}) + 1$,
	    $\hspace*{9 mm} sum(D) := sum(D) + \delta$ \\
	    else $D = D + 1, H(D) = \delta, \\
                \hspace*{6 mm}  count(D) = 1, sum(D) = \delta$.
      \end{itemize}
\item Replace $H$ by the average of all activation values that have been
      clustered into this cluster:
	  \[ H(j) := sum(j)/count(j),\; j = 1,2 \ldots, D. \]
\item  Check the accuracy of the network with the activation values $\delta^i$
       at the hidden nodes replaced by $\delta_d$, the activation value
       of the cluster to which the activation value belongs.

\item  If the accuracy falls below the required level, decrease
       $\epsilon$ and  repeat Step 1.
\end{enumerate}

\item  Enumerate the discretized activation values  and compute the 
       network output. 

	Generate perfect rules  that have a perfect cover of
       all the tuples from the hidden node activation values to the output
       values.
        
\item  For the discretized hidden node activation values appeared in 
       the rules found in the above step, enumerate the input values that 
       lead to them, and generate perfect rules.
 
\item  Generate rules that relate the input values and the output
       values by rule substitution based on
	the results of the above two steps.
\end{enumerate}
\caption{Rule extraction algorithm (RX)}
\label{fig:rule_ex}
\vspace*{2mm} 
\begin{tabular}{cc}
\hspace{72 mm} & \\ \hline
\end{tabular}
\end{figure}

Our  rule extracting algorithm is outlined in Figure~\ref{fig:rule_ex}.
The algorithm first discretizes the activation values of hidden nodes into
a manageable number of discrete values without sacrificing the classification
accuracy of the network.  A small set of the discrete activation
values make it possible to  determine both the dependency among the output
values and the hidden node values and the dependency among the hidden node 
activation values and the input values.

From the dependencies, rules can be generated~\cite{Liu95}.
Here we show the process of extracting  rules
from the pruned network in Figure~\ref{fig:prun_nn}
obtained for the classification problem Function 2.


The network has three hidden nodes.  The activation
values of 1000 tuples were discretized. The value of $\epsilon$ was set to
0.6. The results of discretization are shown in the following table.

\begin{table}[hbt]
\begin{center}
\vspace*{2mm}
\begin{tabular}{c | c | l} \hline \hline
Node & No of clusters    &  Cluster activation values \\ \hline\hline
1   &    3    &   (-1, 0, 1)  \\
2   &    2    &   ( 0, 1)     \\
3   &    3    &   (-1, 0.24, 1) \\ \hline
\end{tabular}
\end{center}
\end{table}

The classification accuracy of the network was checked by replacing
the individual activation value with its discretized activation value.
The value of $\epsilon = 0.6$ was sufficiently small to preserve the
accuracy of the neural network and large enough to produce only a small number
of clusters. For the three hidden nodes,
the numbers of discrete activation values (clusters)
are 3,2 and 3, or a total of 18 different outcomes at the two
output nodes are possible. We tabulate the outputs $C_j (1 \le j \le 2)$
of the network 
according to the hidden node activation values $\alpha_m, (1 \le m \le 3)$ 
as follows.

\begin{table}[hbt]
\begin{center}
\begin{tabular}{r r r |   r r } \hline \hline
$\alpha_1$ & $\alpha_2$ & $\alpha_3$ & $C_1$  & $C_2$ \\ \hline \hline
-1 & 1  & -1   &  0.92  & 0.08 \\
-1 & 1  & 1    &  0.00  & 1.00  \\
-1 & 1  & 0.24 &  0.01  & 0.99  \\
-1 & 0  & -1   &  1.00  & 0.00  \\
-1 & 0  & 1    &  0.11  & 0.89  \\
-1 & 0  & 0.24 &  0.93  & 0.07  \\
1  & 1  & -1   &  0.00  & 1.00  \\
1  & 1  & 1    &  0.00  & 1.00  \\
1  & 1  & 0.24 &  0.00  & 1.00  \\
1  & 0  & -1    &  0.89  & 0.11  \\ 
1  & 0  & 1     &  0.00  & 1.00  \\
1  & 0  & 0.24  &  0.00  & 1.00  \\
0  & 1  & -1    &  0.18  & 0.82  \\
0  & 1  & 1     &  0.00  & 1.00  \\
0  & 1  & 0.24  &  0.00  & 1.00  \\
0  & 0  & -1    &  1.00  & 0.00  \\
0  & 0  & 1     &  0.00  & 1.00  \\
0  & 0  & 0.24  &  0.18  & 0.82  \\ \hline
\end{tabular}
\end{center}
\end{table}

Following Algorithm RX step 2, 
the predicted outputs of the network are taken to be
$C_1  = 1 $ and $C_2 = 0$   
if the activation values $\alpha_m$'s satisfy one of the following conditions
(since the table 
is small, the rules can be checked manually):

\begin{eqnarray*}
R_{11}: C_1 = 1, C_2 = 0  & \Leftarrow & \alpha_2 = 0, \alpha_3 = -1. \\
R_{12}: C_1 = 1, C_2 = 0  & \Leftarrow & \alpha_1 = -1, \alpha_2 = 1, \alpha_3 = -1. \\
R_{13}: C_1 = 1, C_2 = 0  & \Leftarrow & \alpha_1 = -1, \alpha_2 = 0, \alpha_3 = 0.24. 
\end{eqnarray*}
Otherwise, $C_1 = 0 $ and $C_2 = 1$.   

The activation values of a hidden node are determined by the inputs
connected to it.
In particular, the three activation values of hidden node 1
are determined by 4 inputs, ${\cal I}_1, {\cal I}_{13}, {\cal I}_{15},$
and ${\cal I}_{17}$. The activation values of hidden node 2 are determined by
2 inputs  ${\cal I}_2$ and ${\cal I}_{17}$, and the activation values of hidden
node 3 are determined by  ${\cal I}_4, {\cal I}_{5}, {\cal I}_{13},
{\cal I}_{15}$ and ${\cal I}_{17}$.
Note that only 5 different activation values appear 
in the above three rules. 
Following Algorithm RX step 3,
we obtain rules that show how a hidden node is activated for
the five different activation values at the three hidden nodes:

\vspace*{2mm} 

\begin{tabular}{llcl}
\multicolumn{4}{l}{Hidden node 1:} \\
$R_{21}:$ & $\alpha_1 = -1$ & $\Leftarrow$ & ${\cal I}_{13} = 1$\\
$R_{22}:$ &  $\alpha_1 = -1$ & $\Leftarrow$ & ${\cal I}_1 =
	{\cal I}_{13} = {\cal I}_{15} = 0,$\\
          &                   &  &     $ {\cal I}_{17} = 1$  \\
          &                   &  &     \\
\multicolumn{4}{l}{Hidden node 2:} \\
$R_{23}:$ & $\alpha_2 = 1 $ & $\Leftarrow$ & ${\cal I}_{2} = 1$ \\
$R_{24}:$ & $\alpha_2 = 1 $ & $\Leftarrow$ & ${\cal I}_{17} = 1$ \\
$R_{25}:$ & $\alpha_2 = 0 $ & $\Leftarrow$ & ${\cal I}_2 = {\cal I}_{17} = 0$ \\
 & &                   \\
\multicolumn{4}{l}{Hidden node 3:} \\
$R_{26}:$ & $\alpha_3 = -1$ & $\Leftarrow$ & ${\cal I}_{13} = 0$ \\
$R_{27}:$ & $\alpha_3 = -1$ & $\Leftarrow$ & ${\cal I}_5 = {\cal I}_{15}  = 1$ \\
$R_{28}:$ & $\alpha_3 = 0.24$ & $\Leftarrow$ & ${\cal I}_4 = {\cal I}_{13} = 1,\;
                              {\cal I}_{17} = 0$ \\
$R_{29}:$ & $\alpha_3 = 0.24$ & $\Leftarrow$ & ${\cal I}_{5} = 0,\; {\cal I}_{13} =
                              {\cal I}_{15} = 1$ \\ 
\end{tabular}

\vspace*{2mm}
With all the intermediate rules obtained above, we can derive the
classification rules as in Algorithm RX step 4. For example, 
substituting rule $R_{11}$ with rules $R_{25}, R_{26},$ 
and $R_{27}$, we have the following two rules in terms of the
original inputs:
\begin{eqnarray*}
R 1: C_1 = 1, C_2 = 0  & \Leftarrow & {\cal I}_2 = {\cal I}_{17} = 0, {\cal I}_{13} = 0 \\
R_1': C_1 = 1, C_2 = 0  & \Leftarrow  &
{\cal I}_2 = {\cal I}_{17} = 0, {\cal I}_5 = {\cal I}_{15}  = 1
\end{eqnarray*}

Recall that the input values of ${\cal I}_{14} $ to ${\cal I}_{19} $
represent coded age groups where  
${\cal I}_{15} = 1$ if $age$ is in [60, 80) and 
${\cal I}_{17} = 1$ if $age$ is in [20, 40). 
Therefore rule $R_1'$ in fact can never be satisfied by any tuple,
hence redundant.

Similarly, replacing rule $R_{12}$ with $R_{21}, R_{22}$, $R_{23}, R_{24}, 
R_{26}$ and $R_{27}$, we have the following two rules:
\begin{eqnarray*}
R 2: C_1 = 1, C_2 = 0  & \Leftarrow & {\cal I}_5 = {\cal I}_{13} = {\cal I}_{15} = 1. \\
R 3: C_1 = 1, C_2 = 0  & \Leftarrow & {\cal I}_1 = {\cal I}_{13} = {\cal I}_{15} = 0,  {\cal I}_{17} = 1. \\
\end{eqnarray*}

Substituting $R_{13}$ with $R_{21}$, $R_{22}$, $R_{25}$, $R_{28}$ and $R_{29}$,
we have another rule:
\begin{eqnarray*}
R 4: C_1 = 1, C_2 = 0   &  \Leftarrow  &  {\cal I}_2 = {\cal I}_{17} = 0, {\cal I}_4 = {\cal I}_{13} = 1. 
\end{eqnarray*}

It is now trivial to obtain the rules in terms of the original attributes. 
Conditions of the rules after substitution can be rewritten in terms of 
the original
attributes and classification problem as shown in Figure~\ref{fig:rule-f2}.

\begin{myfigure}
\begin{tabular}{l l}
    Rule 1.&If ({\bf salary} $<$ 100000)
	    $\wedge$ ({\bf commission} $=$ 0) $\wedge$ 
	    ({\bf age} $\leq$ 40), then Group A.  \\
    Rule 2.&If ({\bf salary} $\geq$ 25000)
	    $\wedge$ ({\bf commission} $>$ 0) $\wedge$ 
	    ({\bf age} $\geq$ 60), then Group A.  \\
    Rule 3.&If ({\bf salary} $<$ 125000) $\wedge$ 
	    ({\bf commission} = 0) $\wedge$ (40 $\leq$ {\bf age} $\leq$ 60),
	    then Group A.  \\
    Rule 4.&If (50000 $\leq$ {\bf salary} $<$ 100000) $\wedge$
	    ({\bf age } $<$ 40), then Group A.  \\
    Default&Rule. Group B.
\end{tabular}\vspace*{0mm}\\
\caption{Rules generated by NeuroRule for Function 2.}
\label{fig:rule-f2}
\end{myfigure}

Given the fact that $salary \geq 75000 \Leftrightarrow
commission = 0$, the above four rules obtained by the pruned network
are identical to the classification Function 2.

\subsection{Hidden node splitting and creation of a subnetwork}

After network pruning and activation value discretization,
rules can be extracted by examining the possible combinations 
in the network outputs as shown in the previous section.
However, when there are still too many connections between a hidden node 
and input nodes, it is not trivial to extract rules, 
even if we can, the rules may not be easy to understand.
To address the problem, 
a three layer feedforward subnetwork can be employed to simplify 
rule extraction for the hidden node.
The number of output nodes of this subnetwork is 
the number of discrete values of the hidden node, while the 
input nodes are those connected to the hidden node in the original network.
Tuples in the training set  are grouped according to their discretized 
activation values.
Given $d$ discrete activation values $D_{1}, D_{2},\ldots, D_d$,
all training tuples  with activation values equal to $D_{j}$ are given a 
$d$-dimensional target value of all zeros expect for one 1 in position $j$.
A new hidden layer is introduced for this subnetwork.
This subnetwork is trained and pruned
in the same ways as is the original network.
The rule extracting process is applied for the subnetwork to obtain
the rules describing the input and the discretized activation values.

This process is applied recursively to those hidden nodes with too many
input links until the number of connection is small
enough or the new subnetwork
cannot simplify the connections between the inputs and the hidden node
at the higher level. 
For most problems that we have solved, this step is not necessary.
One problem where this step is required by the algorithm
is for a genetic classification problem with 60 attributes. The details 
of the experiment can be found in~\cite{Rudy-spli95}.

\section {Preliminary experimental results}

Unlike the pattern classification research in the AI community where a set of
classic problems have been studied by a large number of researchers, fewer
well documented benchmark problems are available for data mining.  
In this section, we report the experimental results of applying the approach 
described in the previous sections to the data mining problem defined 
in~\cite{Agra-etal93}.  As mentioned earlier, the database
tuples consisted of nine
attributes (See Table~\ref{tab:desc}).  Ten classification functions 
of Agrawal et al.~\cite{Agra-etal93} were
used to generate classification problems with different complexities.  
The training set consisted of 1000 tuples and the testing data sets had
1000 tuples. Efforts were made to generate the data sets as described
in the original functions.
Among 10 functions described, we found that functions 8 and 10 produced
highly skewed data that made classification not meaningful.
We will only discuss functions other than these two.
To assess our approach, we compare the results with that of C4.5,
a decision tree-based classifier~\cite{Quin93}.

\subsection{Classification accuracy}

The following table reports the classification accuracy using
both our system and C4.5 for eight functions.
Here, classification accuracy is defined as
\begin{equation}
accuracy~=~\frac{no~tuples~correctly~classified}{total~number~of~tuples}
\end{equation}

\begin{table}[hbt]
\begin{center}
\vspace*{2mm} 
\begin{tabular}{| c || c c || c c |} \hline \hline
\multicolumn{1}{| c || } {Func.} & \multicolumn{2}{c||} {Pruned Networks}
                              & \multicolumn{2}{ c |} {C4.5} \\ \cline{2-5}
\multicolumn{1}{| c ||}  {no   } & \multicolumn{1}{|c ||} {Training} & \multicolumn{1}{c||}{Testing} &
  \multicolumn{1}{|c ||} {Training}& \multicolumn{1}{c |} {Testing}\\ \hline
 1 &   98.1  &100.0 & 98.3 &100.0 \\
 2 &   96.3  &100.0 & 98.7 & 96.0 \\ 
 3 &   98.5  &100.0 & 99.5 & 99.1 \\
 4 &   90.6  & 92.9 & 94.0 & 89.7 \\
 5 &   90.4  & 93.1 & 96.8 & 94.4 \\
 6 &   90.1  & 90.9 & 94.0 & 91.7 \\
 7 &   91.9  & 91.4 & 98.1 & 93.6 \\
 9 &   90.1  & 90.9 & 94.4 & 91.8 \\ \hline
\end{tabular}
\end{center}
\end{table}

From the table we can see that the classification accuracy of the neural 
network based approach and C4.5 is comparable.  
In fact, the network obtained after the training phase has
higher accuracy than what listed here, which is mainly determined by  
the threshold set for the network pruning phase.  In our experiments,
it is set to 90\%. That is, a network will be pruned until 
further pruning will cause the accuracy to fall below this threshold.  
For applications where high 
classification accuracy is desired, the threshold can be set higher so that
less nodes and links will be pruned.   Of course,
this may lead to more complex classification rules.
Tradeoff between the accuracy and the complexity of the classification
rule set is one of the design issues.

\subsection{Rules extracted}

Here we present some of the classification rules extracted from
our experiments.

For simple classification functions, the rules extracted are exactly the
same as the classification functions.  These include functions 1, 2 and 3.
One interesting example is Function 2.  The detailed process of finding
the classification rules is described as an example in Section 2 and 3.
The resulting rules are the same as the original functions.
As reported by Agrawal {\em et al.}~\cite{Agra-etal93}, 
ID3 generated a relatively
large number of strings for Function 2 when the decision tree is built.
We observed similar results when C4.5rules was used (a member of ID3).
C4.5rules generated  18 rules.  Among the 18 rules, 8 rules
define the conditions for Group A.  Another 10 rules define Group B.  
Tuples that do not satisfy the conditions specified are classified as default
class, Group B.  Figure~\ref{fig:c45-f2} shows the rules that define
tuples to be a member of Group A.

\begin{myfigure}
\begin{tabular}{l l}  
Rule 16: & ({\bf salary} $>$ 45910) $\wedge$ 
	   ({\bf commission} $>$ 0) $\wedge$
	   ({\bf age} $>$ 59) \\
Rule 10: & (51638 $<$  {\bf salary} $\leq$ 98469) $\wedge $
	   ({\bf age} age $\leq$ 39) \\
Rule 13: & ({\bf salary} $\leq$ 98469) $\wedge$ 
	   ({\bf commission} $\leq$ 0) $\wedge$ 
	   ({\bf age} $\leq$ 60) \\
Rule 6:  & (26812 $<$  {\bf salary} $\leq$ 45910) $\wedge$ 
	   ({\bf age} $>$ 61) \\
Rule 20: & (98469 $<$  {\bf salary} $\leq$ 121461) $\wedge$
	   (39 $<$ {\bf age} $\leq$ 57)  \\
Rule 7:  & (45910 $<$  {\bf salary} $\leq$ 98469) $\wedge$ 
	   ({\bf commission} $\leq$ 51486) $\wedge$  
	   ({\bf age} $\leq$ 39) $\wedge $ 
	   ({\bf hval} $\leq$ 705560)   \\
Rule 26: & (125706 $<$ {\bf salary} $\leq$ 127088) $\wedge$ 
	   ({\bf age} $\leq$ 51)  \\
Rule 4:  & (23873 {\bf salary} $\leq$ 26812) $\wedge$ 
	   ({\bf age} $>$ 61) $\wedge$
	   ({\bf loan} $>$ 237756)  \\
\end{tabular}  
\caption{Group A rules generated by C4.5rules for Function 2.}
\label{fig:c45-f2}
\end{myfigure}

By comparing the rules generated by C4.5rules
(Figure~\ref{fig:c45-f2}) with the rules generated by NeuroRule in Figure 4, it is obvious
that our approach generates better rules in the sense that they are more
compact, which makes the verification and application of the rules
much easier.
 
Functions 4 and 5 are another two functions for which ID3 generates a large
number of strings. ${\cal CDP}$~\cite{Agra-etal93} 
also generates a relatively large number of strings than for other functions.
The original classification function 4, the 
rule sets that define Group A tuples extracted using NeuroRule and C4.5, 
respectively are shown in Figure~\ref{fig:func_4}.

\begin{myfigure}
\vspace{2mm}
{\bf (a) Original classification rules defining Group A tuples}
\vspace{2mm}
\begin{tabular}{l l}
Group A: &~$ (({\bf age} < 40) \wedge $ \\
       &~$((({\bf elevel} \in [0..1]) ? (25K \le {\bf salary} \le 75K)) : (50K \le {\bf salary} \le 100K)))) \vee $ \\
       &~$((40 \le {\bf age} < 60) \wedge $ \\
       &~$((({\bf elevel} \in [1..3]) ? (50K \le {\bf salary} \le 100K)) : (75K \le {\bf salary} \le 125K)))) \vee  $ \\
       &~$(({\bf age} \geq 60) \wedge $ \\
       &~$((({\bf elevel} \in [2..4]) ? (50K \le {\bf salary} \le 100K)) : (25K \le {\bf salary} \le 75K))))$ \\
\end{tabular}

\vspace{2mm}
{\bf (b) Rules generated by NeuroRule}
\vspace{2mm}
\begin{tabular}{l l}
$R 1$: & if (40 $\leq$ {\bf age} $<$ 60) $\wedge$ 
           ({\bf elevel} $ \le $ 1) $\wedge$ 
           (75K  $\leq$ {\bf salary}  $<$100K) 
           then Group A  \\
$R 2$: & if ( {\bf age} $<$60) $\wedge$ ({\bf elevel} $\geq$ 2)
              $\wedge$ (50K $ \le $ {\bf salary} $<$100K) 
           then Group A  \\
$R 3$: & if ({\bf age} $<$60)  $\wedge$
              ({\bf elevel} $ \le $ 1) $\wedge$
              (50K $\leq$ {\bf salary} $<$ 75K ) 
           then Group A  \\
$R 4$: & if ({\bf age} $ \geq$ 60) $\wedge$ 
              ( {\bf elevel}  $\leq$ 1) $\wedge$ 
              ({\bf salary} $<$75K) 
           then Group A  \\
$R 5$: & if ({\bf age} $ \geq$ 60) $\wedge$ 
              ({\bf elevel} $ \geq$ 2) $\wedge$ 
              (50K $\leq$ {\bf salary} $<$ 100K) 
           then Group A
\end{tabular}

\vspace{2mm}
{\bf (C) Rules generated by C4.5rules }
\vspace{2mm}
\begin{tabular}{l l}
Rule 30: & ({\bf elevel} $=$ 2) $\wedge$
	   (50762 $<$ {\bf salary} $\leq$ 98490)   \\
Rule 25: & ({\bf elevel} $=$ 3) $\wedge$ 
	   (48632 $<$  {\bf salary} $\leq$  98490)  \\
Rule 23: & ({\bf elevel} $=$ 4) $\wedge$
	   (60357   $<$ {\bf salary} $\leq$ 98490)  \\
Rule 32: & (33 $<$ {\bf age} $\leq$ 60) $\wedge$
	   (48632 $<$ {\bf salary} $\leq$ 98490)$\wedge$  
	   ({\bf elevel} $=$ 1)  \\
Rule 57: & ({\bf age} $>$ 38) $\wedge$ 
	   (102418 $<$ {\bf salary} $\leq$ 124930 $\wedge$ 
	   ({\bf age} $\leq$ 59) $\wedge$ {\bf elevel} $=$ 4) \\
Rule 37: & ({\bf salary} $>$ 48632) $\wedge$ ({\bf  commission} $>$ 18543)  \\
Rule 14: & ({\bf age} $\leq$ 39) $\wedge$  
	   ({\bf elevel} $=$ 0)  $\wedge$ 
	   ({\bf salary} $\leq$ 48632)  \\ 
Rule 16: & ({\bf age} $>$ 59) $\wedge$  
	   ({\bf elevel} $=$ 0) $\wedge$
	   ({\bf salary} $\leq$ 48632)  \\ 
Rule 12: & ({\bf age} $>$ 65) $\wedge$  
	   ({\bf elevel} $=$ 1) $\wedge$
	   ({\bf salary} $\leq$ 48632)  \\ 
Rule 48: & ({\bf car} $=$ 4) $\wedge$ (98490 $<$ {\bf salary} $\leq$ 102418)\\
\end{tabular}
\caption{Classification function 4 and rules extracted.}
\label{fig:func_4}
\end{myfigure}

The five rules extracted by NeuroRule are not exactly the same as the
original function descriptions (Function 4).
To test the rules extracted, the rules were applied to three
test data sets of different sizes, shown in Table~\ref{tab:func_4}.
The column {\em Total \/} is the total number of tuples that are classified
as group A by each rule.  The column {\em Correct \/} is the percentage
of correctly classified tuples.  E.g., rule $R 1$ classifies all
tuples correctly.  On the other hand,  among 165 tuples that were classified
as Group A by rule $R 2$, 6.1\% of them belong to
Group B, i.e. they were misclassified.

\noindent
\begin{table*}[btp]
\begin{center}
\caption{Accuracy rates of the rules extracted for function 4}
\label{tab:func_4}
\vspace*{2mm}
\begin{tabular}{| l|  c c|  c c|  c c |} \hline \hline
 & \multicolumn{6}{c | } {Test data size} \\ \cline{2-7}
Rule & \multicolumn{2}{c|} {1000}
     & \multicolumn{2}{c|} {5000} & \multicolumn{2}{c |} {10000} \\ \cline{2-7}
     & \multicolumn{1}{c} {\hspace*{2mm} Total \hspace*{2mm}} &
       \multicolumn{1}{c|}{Correct (\%)} &
       \multicolumn{1}{c} {\hspace*{2mm} Total \hspace*{2mm}} &
       \multicolumn{1}{c|}{Correct (\%)} &
       \multicolumn{1}{c} {\hspace*{2mm} Total \hspace*{2mm}} &
       \multicolumn{1}{c | }{Correct (\%)} \\ \hline
$R 1$  &  22  & 100.0   & 111  & 100.0  & 239 & 100.0   \\
$R 2$  & 165  & 93.9   & 753  & 92.6 &1463 & 92.3  \\
$R 3$  &  46  & 82.6   & 247  & 78.4 & 503 & 78.3  \\
$R 4$  &  51  & 82.4   & 305  & 87.9 & 597 & 89.4  \\
$R 5$  &  71  &  100.0   & 385  & 100.0  & 802 & 100.0   \\ \hline
\hline \end{tabular}
\end{center}
\end{table*}

From Table~\ref{tab:func_4} , we can see that two of the rules extracted
classify the tuples correctly without errors.  They are exactly the
same as parts of the original function definition.  Because the accuracy of
the pruned network is not 100\%, other rules extracted are not the same
as the original ones.  However, the rule extracting phase preserves 
the classification accuracy of the pruned network.  It is
expected that, with higher accuracy of the network, the accuracy of the
extracted rules will be also improved.   

When the same training data set was used as the input of 
C4.5rules,  twenty rules were generated among which 10 rules define
the conditions of Group A (Figure~\ref{fig:func_4}).  
Again, we can see that NeuroRule generates 
better rules than C4.5rules.  Furthermore, rules generated by NeuroRule 
only reference those attributes appeared in the original classification
functions.  C4.5rules in fact picked some attributes, e.g. {\em car \/},
that does not appear in the original function.

\section {Conclusion}

In this paper we reported NeuroRule, a connectionist approach to mining
classification rules from given databases.  The approach consists
of three phases: (1) training a neural network that correctly classifies
tuples in the given training data set to a desired accuracy; (2)
pruning the network while maintaining the classification accuracy;
and (3) extracting explicit rules from the pruned network.
The proposed approach was applied to a set of classification problems.
The results of applying it to a data mining problem defined 
in~\cite{Agra-etal93} was discussed in detail.  The results
indicate that, using the proposed approach, high quality rules can 
be discovered from the given ten data sets.
While considerable work on using neural networks for classification
has been reported,  none of them can generate rules with
the quality comparable to those generated by NeuroRule.

The work reported here is our first attempt to apply the connectionist
approach to data mining.   A number of related issues are to be
further studied.  One of the issues is to reduce the training time of
neural networks.  Although we have been improving the speed of network
training by developing fast algorithms, the time required for NeuroRule is
still longer than the time needed by the symbolic approach, such as C4.5.
As the long initial training time of a network may be tolerable, 
incremental training and rule extraction during the life time of an 
application database can be useful.
With incremental training that requires less time, 
the accuracy of rules extracted can be improved along with the change of 
database contents.  

\bibliographystyle{plain}
\bibliography{/home/staff2/liuh/BIB/huan}

\end{document}